\documentclass[sigconf]{acmart}
\usepackage{multirow}
\usepackage{graphicx}
\usepackage{subcaption}
\usepackage{float}
\AtBeginDocument{%
  \providecommand\BibTeX{{%
    \normalfont B\kern-0.5em{\scshape i\kern-0.25em b}\kern-0.8em\TeX}}}

\setcopyright{acmcopyright}
\copyrightyear{2023}
\acmYear{2023}
\acmDOI{XXXXXXX.XXXXXXX}

\begin{document}

\title{Seeing and hearing what has not been said}
\subtitle{A multimodal client behavior classifier in Motivational Interviewing with interpretable fusion}


\author{Lucie Galland}
\orcid{0000-0003-4682-6011}
\affiliation{
 \institution{CNRS - ISIR, Sorbonne University}
   \city{Paris}
   \country{France}}
 \email{lucie.galland@isir.upmc.fr}

 \author{Catherine Pelachaud}
 \orcid{0000-0003-1008-0799}
 \affiliation{
   \institution{CNRS - ISIR, Sorbonne University}
   \city{Paris}
   \country{France}}
 \email{catherine.pelachaud@upmc.fr}

\author{Florian Pecune}
 \orcid{0000-0002-3235-2575}
 \affiliation{
   \institution{Bordeaux University}
   \city{Bordeaux}
   \country{France}}
 \email{fpecune@u-bordeaux.fr}

\renewcommand{\shortauthors}{Galland, et al.}

\begin{abstract}
Motivational Interviewing (MI) is an approach to therapy that emphasizes collaboration and encourages behavioral change. To evaluate the quality of an MI conversation, client utterances can be classified using the MISC code as either change talk, sustain talk, or follow/neutral talk. The proportion of change talk in a MI conversation is positively correlated with therapy outcomes, making accurate classification of client utterances essential. 

In this paper, we present a classifier that accurately distinguishes between the three MISC classes (change talk, sustain talk, and follow/neutral talk) leveraging multimodal features such as text, prosody, facial expressivity, and body expressivity. To train our model, we perform annotations on the publicly available AnnoMI dataset to collect multimodal information, including text, audio, facial expressivity, and body expressivity. Furthermore, we identify the most important modalities in the decision-making process, providing valuable insights into the interplay of different modalities during a MI conversation.
\end{abstract}



\begin{teaserfigure}
    \centering
    \includegraphics[width=\textwidth]{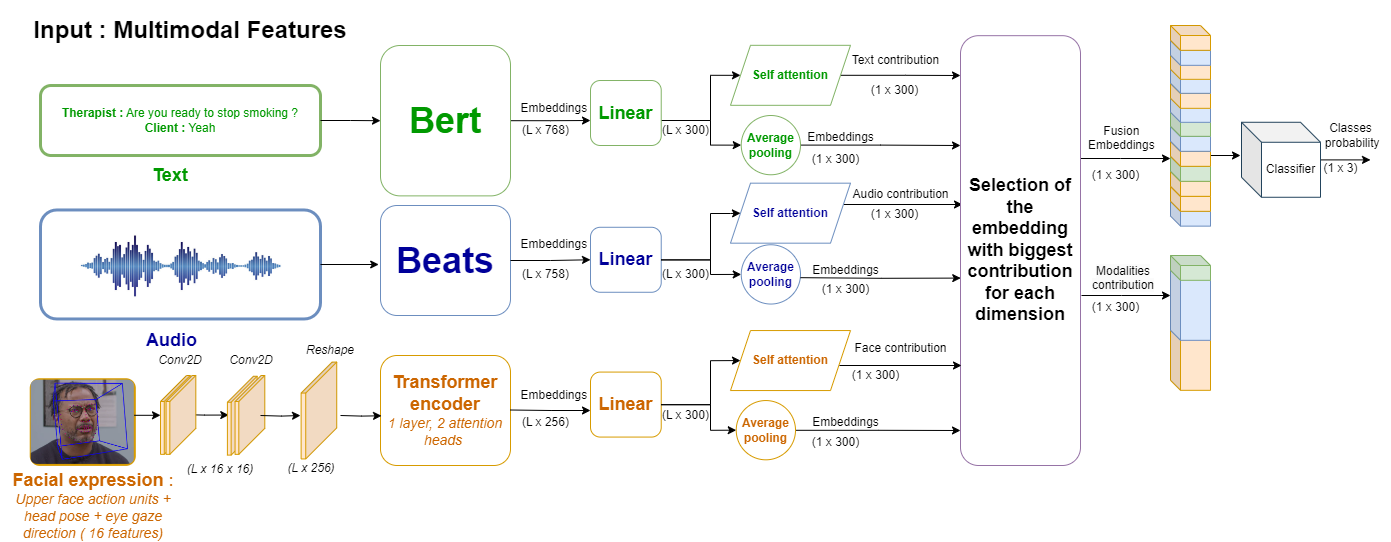}
    \caption{MALEFIC model architecture\\ \tiny{Modality Attentive Late Embracenet Fusion with Interpretable Modality Contribution (MALEFIC)}}
    \label{fig:architecture}
\end{teaserfigure}


\maketitle

\section{Introduction}

Motivational Interviewing (MI) is an approach to therapy that emphasizes collaboration and encourages behavioral change.
During Motivational Interviews, therapists rely on a set of strategies to guide clients toward expressing motivation toward change \cite{miller2012motivational}. Assessment of the quality of the therapy interaction is classically done by annotating therapist's and client's behaviors. To this intent, various annotations schema have been developed such as the Motivational Interviewing Skill Code (MISC) \cite{miller2003manual} that classifies both therapist and client behaviors into three relevant categories:

\begin{itemize}
\item \textbf{Change talk (CT)}: reflecting actions toward behavior change 
\item \textbf{Sustain talk (ST)}: reflecting actions  away from behavior change 
\item \textbf{Follow/Neutral (F/N)}: unrelated to the target behavior
\end{itemize}
This classification of client language is of interest as it is a predictor of the therapy outcome.
Indeed, \cite{magill2014technical} revealed that sustain-talk was associated with poorer treatment results. Furthermore,  \cite{magill2018meta} showed that change talk was linked to reductions in risk behavior during follow-up assessments. This correlation makes MISC a promising tool for studying the efficacy of Motivational Interviewing (MI).

The labeling of client utterances is usually done by training coders to manually encode utterances into these three categories.
However, this process of annotation can be resource-intensive, as it requires trained annotators to carefully review videos. Furthermore, it can not be done in real-time and can not be used in the context of a human-agent dialogue for instance. As a result, there has been growing interest in developing automatic annotation methods for MISC using various modalities and approaches. These efforts aim to streamline the annotation process and reduce the time and resources required for the analysis.

In this paper, we continue these efforts by presenting a classifier that can distinguish automatically between the three MISC classes. This classifier is based on multimodal features of face-to-face conversations, including (spoken) text, prosody, facial expressivity, and body expressivity. Our classifier is designed to be interpretable, meaning that it is possible to identify the modality that was most important in its decision-making process.

In the remaining of the paper, we first present the data we used to train our MISC classifier, then we present our modality attentive fusion architecture. We explore the performance of different models and compare our results with existing work. Finally, we present a way to interpret the results of the classification to shed a light on the contribution of modalities in the classification process.

\section{Related Work}

The correlation between MISC codes and therapy outcomes has motivated several studies to develop their own classification systems for client language, categorizing it as change talk, sustain talk, or follow neutral. These studies use various modalities as inputs.


Text-based modalities have been widely investigated in the context of MISC annotation on different temporal levels. For example, \cite{howes2013using} used topic modeling to predict therapy outcomes at the session level, while \cite{huang2018modeling} incorporated topic angles and session timing (beginning or end) to predict MISC codes at the utterance level. In their work, an utterance represents a turn by either the client or the therapist. More recent advances have been made using deep learning-based approaches, such as those presented in \cite{ewbank2021understanding}, which leveraged word-level features, and in \cite{chen2021feature}, which incorporated additional utterance-level features like Linguistic Inquiry and Word Count (LIWC) for improved annotation accuracy. In the latter work, utterances were segmented after a pause of at least two seconds. While these advancements highlight the ongoing exploration of various feature sets and modalities in the automatic annotation of MISC codes, they also expose a variety of ways to decide the level used for coding as well as the specification of an utterance.

Text is not the only modality that can convey the nuances of change talk. Several studies have incorporated prosody or acoustic features to improve MISC classification. For instance,  \cite{aswamenakul2018multimodal} combined acoustic features with linguistic features to slightly improve the accuracy of change talk detection. Deep learning methods such as Long Short-Term Memory (LSTM) \cite{singla2018using}  has also been employed to predict change talk using both text and audio modalities. In this work, the addition of the audio modality improves the prediction score. More recently, such classification was performed using Transformers \cite{tavabi2020multimodal}. The use of audio generates a loss in performance that can be explained by the low quality of the recordings.

In addition to acoustic cues, other social signals such as laughter have been explored.  \cite{gupta2014predicting} demonstrated that adding laughter as input improved the accuracy of change talk prediction compared to text alone. Furthermore, non-verbal cues such as facial Action Units have been utilized as predictors for change talk, as shown in  \cite{nakano2022detecting} which resulted  in improving the prediction.

While the text remains a commonly studied modality, incorporating prosody, non-verbal, and other multimodal information alongside text has shown promising potential for improving the accuracy and robustness of MISC annotation and prediction tasks.

Although using different modalities can improve classifier performance, one limitation of the above works is that they rely on at most two modalities at a time. Furthermore, understanding the contribution of each modality to the decision process remains a challenge. Only \cite{singla2018using} addressed this by examining attention weights of the fusion layer, revealing that prosody information have more influence at the end of utterances.

To overcome these limitations, the main contributions of our work include:
\begin{itemize}
    \item Developing a MISC classifier using 3 different modalities: text, prosody, and nonverbal behavior
    \item Developing a classifier that identifies the specific modalities that played a key role in the decision-making process. This feature enables practitioners to determine why the classifier made a particular decision.
\end{itemize}

    

\section{Data}

Motivational interviewing data that could be used to train a MISC classifier is difficult to find due to the sensitive nature of the discussed topics. Most of the existing corpora are either private for medical reasons \cite{borsari2012addressing,carey2009computer} or owned privately and payable. Because of this, most studies need to collect a new dataset first and models can not be compared. For instance, \cite{nakano2022detecting} collected their own non public corpus over Zoom and developed a classifier on the resulting corpus. However, Two corpora of MI conversations have recently been published and are publicly available. The High Low-quality MI dataset \cite{perez2019makes} is composed of 249 videos of MI annotations available on YouTube. Some errors remain in the automatic transcription of the videos and even though MISC annotations have been performed, they are not currently available. The second public corpus is AnnoMI \cite{wu2022anno}, a corpus of MI conversations transcribed and annotated with MISC with publicly available annotations. These datasets do not provide multimodal annotations.
\subsection{AnnoMI corpus}

In our work, we rely on the AnnoMI dataset \cite{wu2022anno} to train ou MISC classifier. AnnoMI is a publicly available dataset of MI videos of 7 minutes on average that have been annotated by 133 experts. The videos are designed as a demonstration of either high or low-quality therapy.  Each video is transcribed and each utterance is annotated in term of primary
therapist behavior (question, reflection, therapist input, and others) and client talk type (neutral, change, sustain) using MISC.
In this work, we are interested in the client side of MISC. A client utterance can be annotated into three categories: Change Talk (CT), Sustain Talk (ST), or Follow/Neutral (F/N). An utterance classified as CT conveys movement towards the behavior of change while ST conveys a movement away from the behavior of change. A F/N utterance does not indicate a preference towards or against change.
The data is annotated by MI practitioners into these 3 classes with 0.9 inter-annotators agreement.

From this corpus we use 121 videos: 3 videos were removed because of outdated URLs and 9 were removed for the poor quality of the video stream. The original transcriptions of the AnnoMI dataset are separated into utterances where a new utterance starts every time a new interlocutor is speaking, only the timestamp of the start of each of these utterances is provided.

\subsection{Dataset preprocessing}
In this paper, we take advantage of the publicly available videos of AnnoMI to train a classifier that predicts client's MISC category relying on multimodal behavior. Multimodality gives valuable insights for various tasks such as sentiment analysis \cite{zadeh2017tensor}. Moreover \cite{pampouchidou2017automatic} shows that visual cues such as facial Action Unit occurrences, head pose, eye gaze, and body gestures can be a sign of depression. Therefore in this paper, we study multiple modalities such as (spoken) text, audio (prosody), and facial and body expressivity.

\begin{figure}
    \centering
    \includegraphics[width=\linewidth]{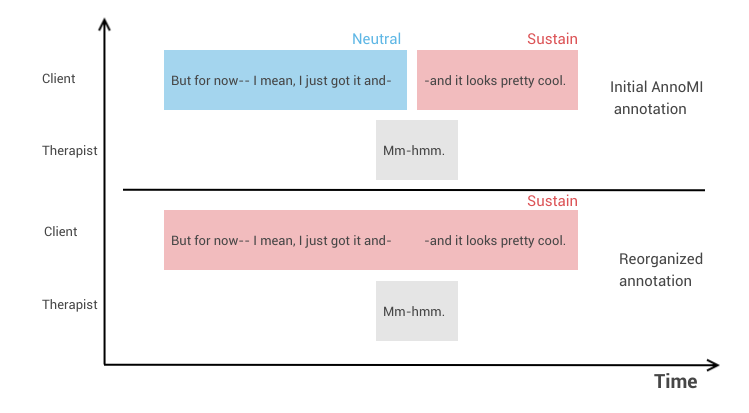}
    \caption{Example of transcript reorganization}
    \label{fig:transcript_reorganization}
\end{figure}

\paragraph{Text}
\label{sec:Text-preprocessing}
In the original AnnoMI transcriptions, sentences were cut into two utterances whenever a listener's backchannel occurred during their production. However, backchannels are not aimed to take the speaking turn. In our model, backchannels are removed from the original transcript and utterances are reorganized to recreate sentences corresponding to speaking turns. We updated the MISC coding whenever utterances of the same sentence received different labels in the original AnnoMI annotation. The only conflicts involved utterances annotated as neutral and change or as neutral and sustain. The resulting sentence is coded as change, respectively sustain. They were no change / sustain conflicts. We illustrate our changes in the Fig.\ref{fig:transcript_reorganization}.

\paragraph{Facial expressivity}

The facial expressivity is extracted using OpenFace \cite{baltrusaitis2018openface}.
As the performance of the OpenFace model is significantly better on videos containing only one face, 
we produce two new videos from the original ones: 
one with the therapist only, and one with the patient only.
In most cases, the camera focuses mainly on the person talking, leaving out of focus the other interlocutor.
Yet, speaking makes the detection of mouth-related action units by OpenFace noisy. 
Therefore, we extract the action units of the upper face (AU 1 2 4 5 6 7 9 and 45). OpenFace is also applied to extract gaze angles and head positions and rotations. 
The action units are smoothed using a median filter with a kernel of size 5 and missing data are interpolated. 

\paragraph{Body expressivity}
Body expressivity can convey information on one's affective state \cite{castellano2007recognising}. Two interesting measures of body expressivity are Amplitude of movement \cite{castellano2007recognising} and Quantity of motion \cite{castellano2008automated}. Amplitude is defined as the width of a movement and Quantity of motion is defined as an approximation of
the amount of detected movement.

Raw body joints position data are extracted using OpenPose \cite{cao2017realtime}. From these raw skeleton data of the client and the therapist, we compute the Amplitude and Quantity of motion for each frame.

The Amplitude is defined as the bounding box around the speaker for a given time frame. It is computed by dividing the length between the two wrists by the height $H$ of the bust in the current framing. Dividing by $H$ accounts for the different sizes in framing.

The quantity of motion QoM is computed following a simplified version of the method described in \cite{castellano2008automated}. Given a silhouette $t$ that moves over $n$ frames, QoM is defined as:
`\begin{equation}
    QoM = Area(Silhouette(t+n)) - Area(Silhouette(t)) 
\end{equation}  We define $Area(Silhouette(t))$ as the bounding box used for the Amplitude and we set n=10 frames.
This simplification is chosen as the interlocutors are seated and the motion is mainly focused on the arms. As the bounding box only takes into account the upper body, the simplification is acceptable.

On both Amplitude and Quantity of motion, missing data are interpolated and a Median filter of size 5 is applied to reduce detection errors from OpenPose.

\subsection{Data distribution}

Similar to other MI datasets \cite{tavabi2020multimodal, nakano2022detecting}, our corpus is unbalanced: the Follow/Neutral class is significantly more prevalent than the Change Talk or Sustain Talk classes (see Table \ref{tab:annomidistrib}). However, our data are more balanced than some previous studies, since we considered speakers' sentences and removed listeners' backchannels.


\begin{table}
\begin{tabular}{l|lll|}
\cline{2-4}
                            & text and audio    & visible face    & visible body  \\ \hline
\multicolumn{1}{|l|}{CT}    & 1279 : 0.24\%     & 1059 :  0.26\%  & 483 : 0.23\%  \\
\multicolumn{1}{|l|}{F/N}   & 3167    : 0. 60\% & 2340 :   0.57\% & 1200 : 0.60\% \\
\multicolumn{1}{|l|}{ST}    & 817 :0.16\%       & 718:  0.17\%    & 353 : 0.17\%  \\ \hline
\multicolumn{1}{|l|}{Total} & 5263 : 100\%      & 4117 : 0.78\%   & 2036:0.39\%   \\ \hline
\end{tabular}

 \caption{AnnoMI distribution}
\label{tab:annomidistrib}
\end{table}
The proportion of each class in the corpus is similar for all modalities, which means that the available modalities are independent of the classes and therefore will not affect the model.

\section{Architecture}
Our MISC classifier relies on the following architecture: each modality of the client input is first prepossessed individually by an adapted network. These encoding networks represent each of the modalities as an embedding vector. The different modalities represented are merged using a modified version of Embracenet \cite{choi2019embracenet}, a fusion architecture that allows missing modalities. We modify Embracenet by adding attention to modalities and call this new architecture MALEFIC (see Section \ref{sec:architecture})
The optimal sizes of the models are determined using a grid search.
\subsection{Modalities pre processing}
\paragraph{Text preprocessing}
The text is preprocessed using a frozen Bert pre-trained model from the HuggingFace library (bert-base-uncased) followed by two linear layers of size 30 interposed with dropout layers, Leaky-Relu activations and one skip connection. We choose to use a frozen Bert model to avoid overfitting.
\paragraph{Text and context preprocessing}
 According to the findings of previous works \cite{nakano2022detecting,tavabi2020multimodal}, we take into account both the therapist's and the client's behaviors. We take as input the previous turn of the therapist, the previous sentences that make up the turn of the client, and the actual client sentence to classify. Each of these sentences is processed sequentially through an un-frozen Bert, and the embeddings obtained from average pooling are concatenated. 
\paragraph{Audio preprocessing}
The Audio modality is preprocessed using the pre-trained Beats model \cite{chen2022beats}. It takes as input the Mel filter bank of the audio and outputs an embedding of size 758.
\paragraph{Facial expressivity preprocessing}
Action Units and head pose values are preprocessed using an encoder composed of two 2-dimensional convolutional layers with 16 filters and a 1-layer Transformer encoder. The encoding of the transformer is then combined to compute an embedding for the entire sequence of size 256.
\paragraph{Body expressivity preprocessing}
Amplitude and Quantity of motion are preprocessed using an encoder composed of 2 convolutional layers and a 1 layer transformer encoder. The encoding of the transformer is then combined to compute an embedding for the entire sequence of size 8.
\subsection{Fusion}
\label{sec:architecture}
The fusion of modalities is achieved using a modified version of Embracenet. This method is useful for handling missing modalities. First, each preprocessing network's output is reduced to the size of the final embedding by a linear layer. Then, Embracenet combines the embeddings by randomly selecting one modality per embedding dimension. In addition, dropout of modality is used during training to prevent over fitting on modalities. During training, modality dropout involves randomly removing available modalities.

This approach enables each preprocessing network to efficiently learn the data structure while also taking advantage of multimodality. Furthermore, it enables us to address missing data in our corpus (namely, the face and body information that are not available for every sentence). In fact, as a result of this training, any missing modality can be easily ignored.

We improve the EmbraceNet architecture by incorporating self-attention. Self-attention is used to determine the significance of a given modality. If a modality is deemed important by the self-attention module, then this modality will be more likely to be selected (see Fig. \ref{fig:architecture}).  


The output of the self-attention layer gives the weight of each modality for each embedding dimension. 
During training, the output of the self-attention layer for a given embedding dimension is used as the probability of selecting each modality. During the evaluation, the selected modality for a given embedding dimension is the modality with the highest probability. We choose to use probabilistic selection during training to avoid over fitting.

We enhance the Embracenet framework with self-attention, as some of the modalities in our problem contribute more to the classification. (for instance, the Text modality has a more substantial classification power than the nonverbal modality, see Tab.\ref{tab:results_single_modality}).

The resulting architecture also estimates the usefulness of each modality, which allows for interpretation (see Section \ref{sec:interpretable})

In the following, we use this architecture that we call : Modality Attentive Late Embracenet Fusion with Interpretable Modality Contribution (MALEFIC), with diffrerent combinations of modalities : Facial and body expressivity; Text and context; Text, context and audio; Text, context and facial expressivity; and Text, context, audio and facial expressivity. For Text and context, we previously took the context into account by concatenating the Bert embeddings of the surrounding sentences. 
Here, we take advantage of our fusion architecture and treat the context as another modality. A self-attention layer will decide whether in this case the client-therapist context is relevant.

\section{Classification results}
To explore the performance of our architecture to predict the MISC classes, we train and evaluate different models using the data described in Section 3. The unbalanced data set is handled using a weighted random sampler.
First, we evaluate the performance of each modality regarding the classification by training different unimodal classifiers. Then, we investigate whether multimodality improves the performance of our best unimodal model. Finally, we compare our results to existing multimodal MISC classification models.

\subsection{Single modality models}
Our first objective is to evaluate which modality allows for the best MISC classification score. 
To that extent, we train different models that take as input a single modality. These models are composed of the preprocessing networks described above, followed by a linear classifier.
The results summarized in Table \ref{tab:results_single_modality} show that the text + context modality appears to be the most efficient. On the other hand, body expressivity has low prediction power. Confidence intervals are calculated using the bootstrap method \cite{efron1994introduction}. Training details are provided below.
\paragraph{Text based model}
The text preprocessing model is trained for 150 epochs with an AdamW optimizer\cite{loshchilov2017decoupled} and a Cosine Aligned scheduler \cite{loshchilov2016sgdr} with a maximum learning rate of $2*10^{-4}$.
\paragraph{Text and context based model}
The text and context preprocessing model is trained for 25 epochs with an AdamW optimizer \cite{loshchilov2017decoupled} and a learning rate of $2*10^{-5}$.
\paragraph{Audio based model}
The audio preprocessing model is trained for 25 epochs with an AdamW optimizer \cite{loshchilov2017decoupled} and a learning rate of $10^{-5}$.
\paragraph{Facial expressivity based model}
The facial expressivity preprocessing model is trained for 150 epochs with an AdamW optimizer \cite{loshchilov2017decoupled} and a One Cycle LR scheduler \cite{smith2017super}with a maximum learning rate of $10^{-4}$.
\paragraph{Body expressivity based model}
The body expressivity preprocessing model is trained for 1500 epochs with an AdamW optimizer \cite{loshchilov2017decoupled} and a learning rate of $5*10^{-5}$.

\begin{table*}
\begin{tabular}{llllll|}
\cline{1-6}
\multicolumn{1}{|l|}{modality :}&Text without context&  Text + context (linear)  & Audio & Facial expressivity & Body expressivity\\ \hline
\multicolumn{1}{|l|}{F1 - CT}   &0.62[0.56,0.68]& \textbf{0.72}[0.66,0.77] & 0.32[0.26,0.39]    &  0.30 [0.23,0.36]     & 0.14[0.05,0.22]     \\
\multicolumn{1}{|l|}{F1 - ST}   &0.63[0.58,0.67]&  \textbf{0.71}[0.67,0.75]& 0.44[0.39,0.5]   &      0.36 [0.31,0.42] &  0.25[0.17,0.35]  \\
\multicolumn{1}{|l|}{F1 - F/N}  &0.79[0.77,0.82]&  \textbf{0.85}[0.83,0.87]  & 0.74[0.71,0.76]    & 0.58 [0.54,0.61]      &   0.67[0.63,0.72]  \\ \hline
\multicolumn{1}{|l|}{F1 - micro }  &0.73[0.70,0.75]& \textbf{0.80}[0.76,0.82]   &  0.62[0.59,0.65]    &  0.46 [0.43,0.49]      & 0.51[0.46,0.55]        \\
\multicolumn{1}{|l|}{F1 - macro }  &0.68[0.65,0.71]&  \textbf{0.76}[0.74,0.79] &   0.51[0.47,0.54] &   0.41[0.38,0.45]         &  0.36[0.31,0.40]     \\\hline
\end{tabular}
\caption{F1 score of single-modality models}
\label{tab:results_single_modality}
 \end{table*}

\subsection{Multimodal models}
Now that we learned more about our unimodal models performance, we investigate whether multimodality could improve the performance of our MISC classification model.
Using the fusion architecture described above, we train several multimodal models. We use a frozen Bert and Beats models to improve training time and avoid over fitting. As a mean of comparaison, we also train the model using text and context linearly from the previous section with a frozen-Bert transformer. 
These multimodal models are trained for 150 epochs with AdamW optimizer \cite{loshchilov2017decoupled} and Cosine Aligned scheduler \cite{loshchilov2016sgdr} with a maximum learning rate of $2*10^{-4}$.
The results are displayed in Table \ref{tab:multimodal_results}.
Because of the low diversity of body expressivity (clients are seated in the videos and do not move much) and the large number of missing data (a quarter of sentences are provided with body expressivity information),
 the addition of body expressivity 
 decreases the accuracy of change talk detection, which is the most important classe. 
 Therefore, in the following, we decide not to use body expressivity in the model.



In all cases, using the MALEFIC architecture improves classification results over the most performant preprocessing network (Text + context linear)
 Particularly, combining text, context, audio, and facial expressivity outperforms all models with frozen Bert and Beats embeddings. Meaning that the combination of visual, vocal, and verbal modalities improves the classification performance. MALEFIC is able to take advantage of the new modalities and to select relevant multimodal information.
 For a MISC classifier, we especially want to be able to classify change talk and avoid classifying change talk as sustain talk and vice versa.  
 The confusion matrix in Tab.\ref{tab:confusionmatrix} shows that our model makes few change talk/sustain talk mistakes.
\begin{table}
    \centering
    \begin{tabular}{@{}cc| ccc|@{}}
\multicolumn{1}{c}{} &\multicolumn{1}{c}{} &\multicolumn{3}{c}{Predicted} \\
\cline{3-5}
\multicolumn{1}{c}{} & 
\multicolumn{1}{c}{} & 
\multicolumn{1}{|c}{ST} & 
\multicolumn{1}{c}{F/N} &\multicolumn{1}{c|}{CT} \\
\cline{2-5}
\multirow[c]{3}{*}{\rotatebox[origin=tr]{90}{Actual}}
&\multicolumn{1}{|l|}{ST}  & 0.65 & 0.29  & 0.06  \\
&\multicolumn{1}{|l|}{F/N}  & 0.07   & 0.79 & 0.14 \\ 
& \multicolumn{1}{|l|}{CT} & 0.04 & 0.27 & 0.69

\\
\cline{2-5}
\end{tabular}
    \caption{Confusion matrix of the model Text+Audio+Face}
    \label{tab:confusionmatrix}
\end{table}
\begin{table*}
\begin{tabular}{|l|llllll|}
\hline
modalities: &\begin{tabular}[c]{@{}l@{}}Text + context\\ (linear)\end{tabular} &\begin{tabular}[c]{@{}l@{}}Text + context\\ (MALEFIC)\end{tabular}  & Face + Body      & Text + Face  & Text + Audio &Text + Audio + Face \\ \hline
F1 - CT    &0.61[0.54,0.66]&0.63[0.57,0.68]&  0.24[0.18,0.31]& 0.64[0.58,0.69]       & \textbf{ 0.65}[0.59,0.70]    &   \textbf{ 0.65}[0.59,0.71]                   \\
F1 - ST    &0.58[0.53,0.63]&0.63[0.58,0.68]&     0.41[0.35,0.47]& 0.60[0.55,0.66]    &  \textbf{0.66}[0.62,0.70]   &    \textbf{0.66}[0.61,0.71]              \\
F1 - F/N  & 0.78[0.75,0.80]&0.80[0.77,0.82]&    0.63[0.60,0.67] & 0.80[0.78,0.83]   & \textbf{0.81}[0.78,0.83]     &    \textbf{ 0.81}[0.77,0.82]            \\ \hline
F1 - micro &0.71[0.68,0.73]&0.73[0.70,0.76]&  0.51[0.47,0.54] & 0.74[0.71,0.76]     &   0.74[0.72,0.77]    &      \textbf{0.76}[0.72,0.77]                \\ 
F1 - macro &0.65[0.62,0.69]&0.69[0.65,0.72]        &  0.43[0.40,0.46] &   0.68[0.65,0.71]                 &\textbf{ 0.71}[0.67,0.73]         &   0.70[0.67,0.73]     \            \\ 
\hline
\end{tabular}

\caption{ F1 score of models trained with frozen Bert and Beats models}
\label{tab:multimodal_results}
\end{table*}

\subsection{Comparison with existing studies}

We compare our results with three existing studies \cite{tavabi2020multimodal,nakano2022detecting,fi15030110}. However, the data set used in these studies is not available, so the conclusion of the comparison should be made with care. The Table \ref{tab:compare} summarizes our comparisons.
\subsubsection{Text based model}
In \cite{wu2022anno}, a Bert model is trained on AnnoMI to predict MISC classes only on the current utterance (text without context). This model is similar to the one we described in section 5.1 and is trained on the same dataset. The only difference with our work is the reorganization of the transcripts performed in Section \ref{sec:Text-preprocessing}. The model in \cite{wu2022anno} reaches a 0.55 F1 macro score, which is significantly lower than the score achieved by our approach (0.68), which uses a similar architecture.

One factor that may explain the performance gap is the preprocessing of the text performed in our approach, as discussed in Section \ref{sec:Text-preprocessing}. By providing full sentences with semantic meaning, our approach is able to capture more nuanced linguistic features, enabling a more accurate classification of MISC classes. These results provide a validation of the effectiveness of our text preprocessing.
\subsubsection{Text and audio-based model}
In \cite{tavabi2020multimodal}, audio and text are used to classify utterances into the 3 MISC classes, change talk, sustain talk, and follow / neutral.
Our approach achieves a significantly higher F1 micro score of 0.62 compared to their score of 0.53, based solely on audio input (see Table \ref{tab:compare}). However, this accuracy gap may be attributed to the poor quality of audio recordings in their corpus, which is not the case in ours.

Moreover, in their approach, adding the audio modality results in a small drop in precision, where, using our fusion method, we are able to slightly improve accuracy by adding the audio modality.
\subsubsection{Text and Facial expressivity based model}
In \cite{nakano2022detecting}, text and facial expressivity (action units, head positions, and eye direction) are used to predict whether an utterance displays change talk or not. They looked at a two-label classification problem when we classify utterances into 3 categories. Their corpus was collected using Zoom, meaning that participants are always facing the camera, whereas our corpus shows a greater variety of body orientations and, therefore, noisier OpenFace outputs. However, we are able to classify change talk significantly better.

In their approach, adding facial expressivity improves the F1 scores on the not change talk class, but does not change the change talk F1 score. Our approach allows us to slightly improve the F1 score on change talk and to produce a higher overall F1 score despite the variety of positions of the clients in the videos and the missing data (when the camera does not show the client's face).

\begin{table*}

\resizebox{\textwidth}{!}{%

\begin{tabular}{|l|lllll|l|llll|llll|llll|}
\hline
modalities                                                                                          & \multicolumn{5}{l|}{Text}                                                                                                                                                                                                                                                                                                                                                                                               & Audio         & \multicolumn{4}{l|}{Text + Audio}                                                                                            & \multicolumn{4}{l|}{Facial expressivity}                                                                     & \multicolumn{4}{l|}{Text + Facial expressivity}                                                              \\ \hline
metric                                                                                              & \multicolumn{1}{l|}{CT}                                                              & \multicolumn{1}{l|}{ST}                                                             & \multicolumn{1}{l|}{F/N}                                                            & \multicolumn{1}{l|}{Micro}                                                          & Macro                                                          & Micro         & \multicolumn{1}{l|}{CT}            & \multicolumn{1}{l|}{ST}            & \multicolumn{1}{l|}{F/N}           & Micro         & \multicolumn{1}{l|}{CT}            & \multicolumn{1}{l|}{ST}   & \multicolumn{1}{l|}{F/N}  & Macro          & \multicolumn{1}{l|}{CT}            & \multicolumn{1}{l|}{ST}   & \multicolumn{1}{l|}{F/N}  & Macro          \\ \hline
 \multicolumn{1}{|l|}{\begin{tabular}[l]{@{}l@{}}MALEFIC*\\\small{Our model}\end{tabular}}                                                                                             & \multicolumn{1}{l|}{\textbf{\begin{tabular}[c]{@{}l@{}}u: 0.62\\ c: 0.72\end{tabular}}} & \multicolumn{1}{l|}{\textbf{\begin{tabular}[c]{@{}l@{}}u: 0.63\\ c: 0.85\end{tabular}}} & \multicolumn{1}{l|}{\begin{tabular}[c]{@{}l@{}}u: 0.79\\ c: 0.71\end{tabular}} & \multicolumn{1}{l|}{\textbf{\begin{tabular}[c]{@{}l@{}}u: 0.73\\ c: 0.80\end{tabular}}} & \begin{tabular}[c]{@{}l@{}}u: 0.68\\c: 0.76\end{tabular} & \textbf{0.62} & \multicolumn{1}{l|}{\textbf{0.65}} & \multicolumn{1}{l|}{\textbf{0.66}} & \multicolumn{1}{l|}{0.80}          & \textbf{0.74} & \multicolumn{1}{l|}{\textbf{0.36}} & \multicolumn{1}{l|}{0.30} & \multicolumn{1}{l|}{0.58} & 0.41           & \multicolumn{1}{l|}{\textbf{0.64}} & \multicolumn{1}{l|}{0.60} & \multicolumn{1}{l|}{0.80} & \textbf{0.74}           \\ \hline
\begin{tabular}[c]{@{}l@{}}Wu, Zixiu, et al*\cite{wu2022anno}\end{tabular}      & \multicolumn{1}{l|}{u: 0.51}                                                          & \multicolumn{1}{l|}{u: 0.39}                                                         & \multicolumn{1}{l|}{u: 0.74}                                                         & \multicolumn{1}{l|}{-}                                                              & u: 0.55                                                         & -             & \multicolumn{1}{l|}{-}             & \multicolumn{1}{l|}{-}             & \multicolumn{1}{l|}{-}             & -             & \multicolumn{1}{l|}{-}             & \multicolumn{1}{l|}{-}    & \multicolumn{1}{l|}{-}    & -              & \multicolumn{1}{l|}{-}             & \multicolumn{1}{l|}{-}    & \multicolumn{1}{l|}{-}    & -              \\ \hline
\begin{tabular}[c]{@{}l@{}}Tavabi et al \cite{tavabi2020multimodal}\end{tabular} & \multicolumn{1}{l|}{-}                                                               & \multicolumn{1}{l|}{-}                                                              & \multicolumn{1}{l|}{-}                                                              & \multicolumn{1}{l|}{\begin{tabular}[c]{@{}l@{}}u: 0.701 \\ c: 0.721\end{tabular}}       & -                                                              & 0.531         & \multicolumn{1}{l|}{0.63}          & \multicolumn{1}{l|}{0.47}          & \multicolumn{1}{l|}{\textbf{0.81}} & 0.714         & \multicolumn{1}{l|}{-}             & \multicolumn{1}{l|}{-}    & \multicolumn{1}{l|}{-}    & -              & \multicolumn{1}{l|}{-}             & \multicolumn{1}{l|}{-}    & \multicolumn{1}{l|}{-}    & -              \\ \hline
Nakanao et al. \cite{nakano2022detecting}                                          & \multicolumn{1}{l|}{\begin{tabular}[c]{@{}l@{}}u: 0.544\\ c: 0.600\end{tabular}}         & \multicolumn{2}{l|}{\begin{tabular}[c]{@{}l@{}}u: 0.874 \\ c: 0.826\end{tabular}}                                                                                             & \multicolumn{1}{l|}{-}                                                              & \begin{tabular}[c]{@{}l@{}}u: 0.709\\ c: 0.666\end{tabular}        & -             & \multicolumn{1}{l|}{-}             & \multicolumn{1}{l|}{-}             & \multicolumn{1}{l|}{-}             & -             & \multicolumn{1}{l|}{0.151}         & \multicolumn{2}{l|}{\textbf{0.836}}                   & \textbf{0.493} & \multicolumn{1}{l|}{0.600}         & \multicolumn{2}{l|}{\textbf{0.873}}                   & \textbf{0.735} \\ \hline
\end{tabular}

}

\caption{Comparison with other studies \small (* = trained using the same corpus, u = without context, c = with context)}
\label{tab:compare}
\end{table*}
\section{Interpretation}
\label{sec:interpretable}
The ability to quantify the contribution of each modality in the classification process is a key advantage of our approach. By utilizing multiple modalities, such as text, prosody and facial expressivity, we can gain a more comprehensive understanding of the client's communication and behavior during an MI conversation.

Identifying which modality is  relevant to the classification of a given sentence can offer valuable insights into the client's state of mind. For example, if facial expressivity or prosody are found to be more influential in the classification process, it may suggest that the client is trying to conceal their true thoughts. 
Several elements of our model offer the bases to draw explanations of the model outputs.
We can name the use of dropout and random selection of embeddings during training allows the final embeddings of each modality to be computed in the same embedding space as the fusion embedding. This ensures that all modalities are represented consistently.

Furthermore, the self-attention layers included in our approach allow the model to dynamically weigh the importance of each modality for each sentence. These layers give a sense of the relevance of each modality not only for each embedding but also for each sentence to be classified. 

In this section, we take advantage of these properties to visualize and quantify the contribution of each modality.
All the following statistics are computed on the part of the validation set where all modalities are available.

\subsection{Overall modality contributions}

To quantify the contribution of each modality within the corpus, we examine the average number of times a modality is selected by the self-attention module over all embedding dimensions. Our analysis reveals the following overall contribution: text (26\%), audio (16\%), face (26\%), previous client sentence in the turn (16\%), and previous therapist turn (16\%). This distribution shows that all modalities are considered by the model with more weight given to the Text and Facial expressivity. These results demonstrate that the model considers all modalities, with a greater weight placed on text and facial expressivity. This aligns with our finding that text is the strongest predictor when taken as a single input (see Table \ref{tab:results_single_modality}). The fact that facial expressivity has a strong weight despite its low predictive powers can be explained in the following sections (see Section \ref{sec:interpre_sentence}).

\subsection{Embedding specialization}
\begin{figure}
    \centering
    \includegraphics[width=\linewidth]{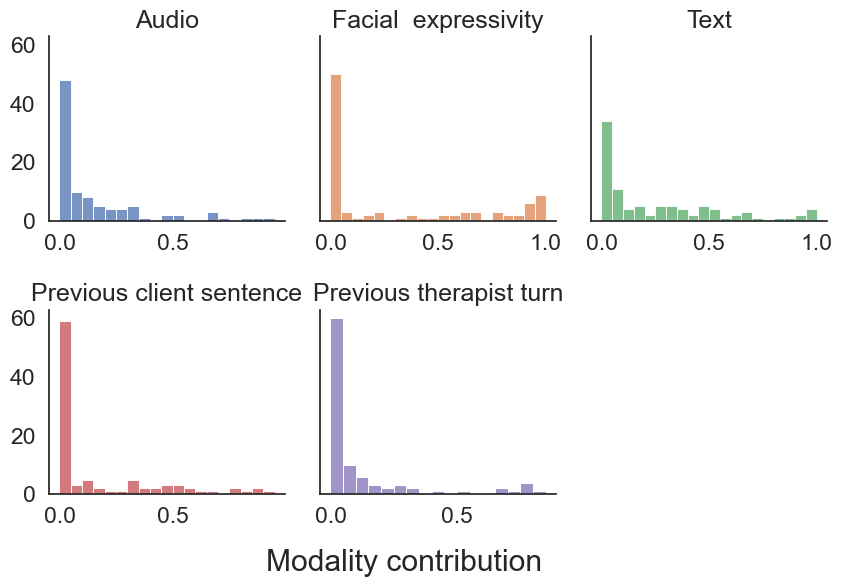}
    \caption{Distribution of modalities contribution for each embedding dimension}
    \label{fig:emb_contribution}
\end{figure}

To understand the role of each embedding dimension, we examine the average number of times a modality was selected for a given embedding dimension. Figure \ref{fig:emb_contribution} shows the distributions of the modality contribution averaged over each embedding dimension. 

This figure shows that some embedding dimensions have a modality contribution of 1 for the text and facial expressivity modalities. This means that this dimension has specialized into a certain modality. This modality will be systematically selected if available. The two modalities that have the greater weight in the overall corpus (text and facial expressivity) are the two modalities with specialized embeddings.  The fact that the dimensions are specialized in the text modality aligns with our finding that the text is the strongest predictor when taken as a single input (see Table \ref{tab:results_single_modality}).

On the other hand, there are, for every modality, some dimensions with a contribution of 0 meaning that this modality is never selected for this dimension.

\subsection{Quantification of modality contribution for each sentence}
\label{sec:interpre_sentence}
\begin{figure}
    \centering
    \includegraphics[width=\linewidth]{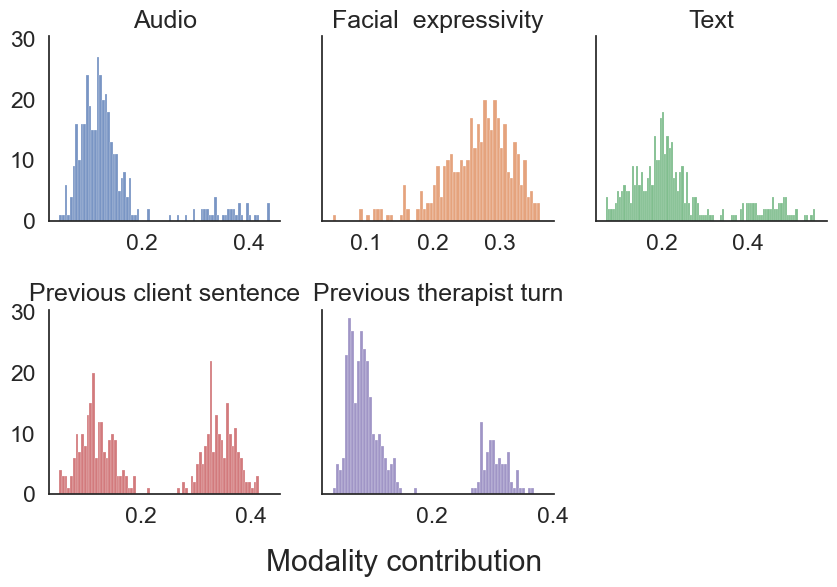}
    \caption{Distribution of modalities contribution for each sentence}
    \label{fig:sentence_contribution}
\end{figure}

\begin{table*}[!ht]
\resizebox{\textwidth}{!}{%
\begin{tabular}{|ll|l|}
\hline
Cluster & Important modality & Context       \\ \hline
1    & Therapist turn  & Therapist: Okay So you were thinking that maybe exposing Lilly naturally to these diseases would be a better choice than using vaccines to help her get stronger? \\
 & & Client: Well, yeah \\\hline

2    & Therapist turn & Therapist: You decided to drink more than you intended because you were disappointed at how the Vikings \\
& & were playing, and when your roommate couldn't give you a ride home, you decided to drive yourself home...                         \\
& &  Client: Yeah, that's, that's exactly how it happened                       \\\hline
3    & Previous client sentence & Therapist: Um, I did wanna talk to you though I'm a little bit concerned looking through his chart \\
& & at how many ear infections he's had recently, and I, I noticed that you had checked the box that someone's ...   \\
          
 & & Client: Well,  It's just me and him, and I do smoke Um, I try really hard not to smoke around him, but I, I've been smoking for 10 years except when I was pregnant with him  \\
 & &                        Client:  But it, everything, it's so stressful being a single mom and, and my having a full-timejob     \\ \hline
4    & Current sentence & Therapist: This what, what, what was  different?   \\
 & &Client: Uh, I don't wanna lose my license  \\
 &  &   Client: You know, I don't, you know, I don't wanna lose my license \\\hline
 5    & Audio &  Therapist: Yeah, it sounds like you'd be willing to do whatever you can to try to prevent that from happening            \\
& & Client : Okay \\\hline

\end{tabular}
}
\caption{Example of transcript for each cluster}
\label{tab:clusterExemple}
\end{table*}
To quantify the contribution of each modality to the classification of a given sentence, we examine the number of dimensions of the fusion embedding that have been selected from this modality for a particular sentence. This provides insights, for a given instance of the client's speech (a sentence), of the amount of information of a modality that is  used to make a decision. Figure \ref{fig:sentence_contribution} shows the distribution of the modality contribution averaged over each sentence. Our analysis indicates that the contribution of each modality is highly dependent on sentences. Specifically, we observed that the distribution of text, audio, and context from both the client and the therapist can be characterized by two Gaussian distributions, indicating that these modalities are more informative for some sentences than for others. 

In contrast, only one Gaussian distribution is visible for facial expressivity, suggesting that this modality is used more consistently across the dataset. This may be because facial expressivity is not a strong predictor for classifying MISC classes. Indeed, because of the use of modality dropout, the model is not able to completely ignore a modality. Therefore, in case of weak predictor, the model has a harder time determining when the modality is useful and takes it into account consistently across the corpus.
This can also explain why facial expressivity has a weight as large as the text modality in the overall contribution and why some embeddings are specialized in this modality. Indeed, the face modality is always selected as the model is not able to detect the sentences where it is really useful and the other modalities are selected only when they are relevant.

To better understand the differences in the sentences that lead to the above results,
we perform a clustering of the contribution of each of the considered modalities using the elbow method and K-means and find five clusters with a silhouette score of 0.96. 

\begin{figure}[H]
    \centering
    \includegraphics[width=\linewidth]{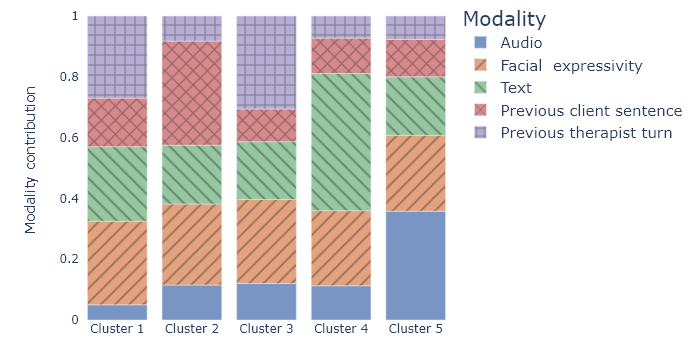}
    \caption{Proportion of modalities contribution within each cluster}
    \label{fig:cluster}
\end{figure}

Sentences can be clustered into groups where the contributions of the modalities are different (see Fig. \ref{fig:cluster}). The five clusters can be interpreted as five types of sentences:
\begin{itemize}
    \item Cluster 1: The text and the context of both, the client and the therapist are relevant: 57\%
    \item Cluster 2: The previous speaking turn of the therapist is relevant: 16\%
    \item Cluster 3: The previous sentences of the client in the speaking turn are relevant: 12\%
    \item Cluster 4: The current sentence is relevant: 9\%
    \item Cluster 5: The audio is relevant: 6\%
\end{itemize}
Table \ref{tab:clusterExemple} shows an example of sentences for each group.

These clusters confirm that facial expressivity contributes consistently across the dataset. Additionally, they demonstrate the importance of considering multiple modalities. By revealing which modality is most relevant for a given sentence, this analysis provides a valuable tool for validating decisions and could be used by the therapist to provide feedback to the client in real-time. It could also be used by a virtual agent acting as the therapist to detect change talk and to use this information for its next dialog move. For example, the agent could explain its decisions by saying something like ``From your tone of voice, it sounds like you are not ready to change.''. As foreseen, the cluster distributions display that text and context are the most important features in most cases.
\subsection{Embedding visualization}

\begin{figure}
\hspace*{\fill}\begin{subfigure}{0.5\linewidth}
\includegraphics[width=\linewidth]{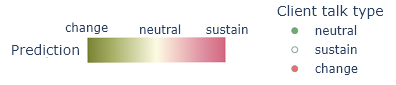}
\end{subfigure}
\medskip
\begin{subfigure}{0.32\linewidth}
\includegraphics[width=\linewidth]{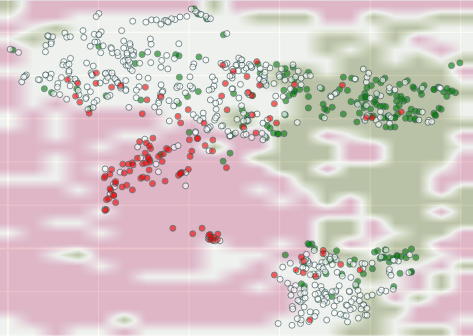}
\caption{Fusion }\label{fig:umap_fusion}
\end{subfigure}\hspace*{\fill}
\begin{subfigure}{0.32\linewidth}
\includegraphics[width=\linewidth]{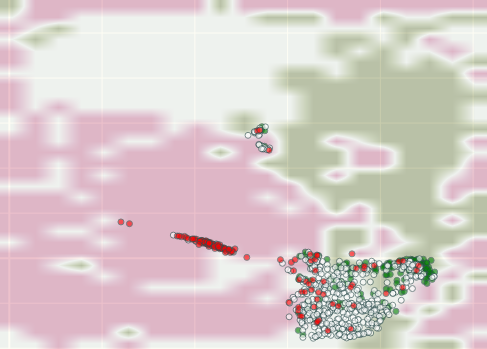}
\caption{Text} \label{fig:umap_text}
\end{subfigure}\hspace*{\fill}
\begin{subfigure}{0.32\linewidth}
\includegraphics[width=\linewidth]{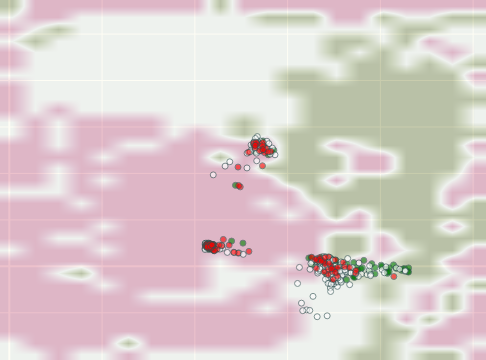}
\caption{Client context} \label{fig:umap_context}
\end{subfigure}
\medskip
\begin{subfigure}{0.32\linewidth}
\includegraphics[width=\linewidth]{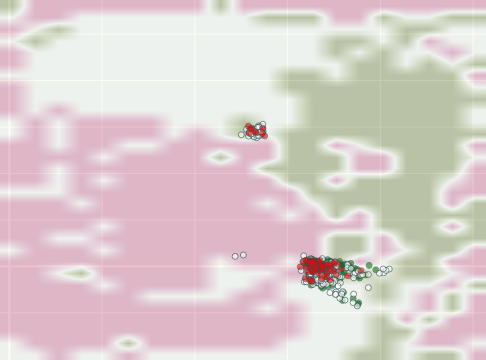}
\caption{Therapist context} \label{fig:umap_therapist}
\end{subfigure}\hspace*{\fill}
\begin{subfigure}{0.32\linewidth}
\includegraphics[width=\linewidth]{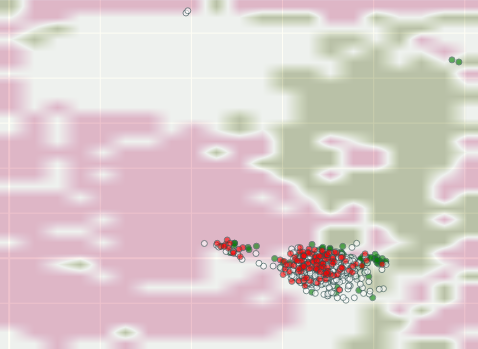}
\caption{Audio} \label{fig:umap_audio}
\end{subfigure}\hspace*{\fill}
\begin{subfigure}{0.32\linewidth}
\includegraphics[width=\linewidth]{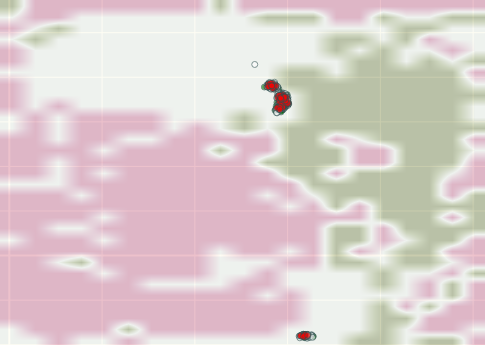}
\caption{Facial expressivity} \label{fig:umap_face}
\end{subfigure}
\caption{Visualization of modalities embeddings with UMAP projection} \label{fig:umap}
\end{figure}

\begin{figure}
\hspace*{\fill}\begin{subfigure}{0.5\linewidth}
\includegraphics[width=\linewidth]{figures/umap_legend.png}
\end{subfigure}
\medskip
\begin{subfigure}{\linewidth}
\includegraphics[width=0.475\linewidth]{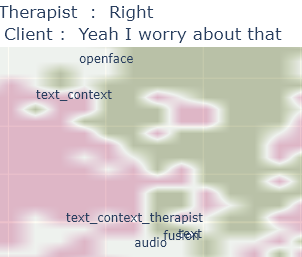}
\hspace*{\fill}
\includegraphics[width=0.475\linewidth]{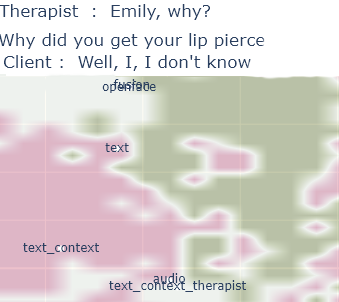}
\caption{Classification improved by multimodality}\label{fig:other_better}
\end{subfigure}
\medskip
\begin{subfigure}{0.475\linewidth}
\includegraphics[width=\linewidth]{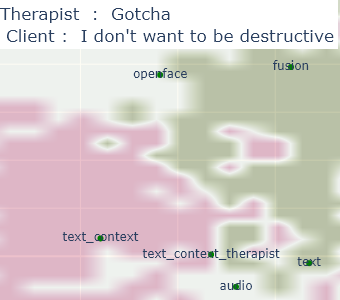}
\caption{Better classification with only text} \label{fig:text_better}
\end{subfigure}
\caption{Examples of sentences representation} \label{fig:umap_sentence}
\end{figure}

The embedding space is visualized using UMAP \cite{mcinnes2018umap}, a framework used for dimensionality reduction that is reversible. Due to its reversible quality, we are able to create a map of the embedding space showing how each embedding point would be classified.
This visualization visible in Figure \ref{fig:umap} allows us to determine how confident the classification is for every modality.
The text is indeed the most expressive modality (see Fig. \ref{fig:umap_text}) and that most of the other modalities are pertinent to accurately classify only in some cases, as seen in the previous sections.
This visualization illustrates also which modalities contributed and in which direction to the classification of each sentence.
Figure \ref{fig:umap_sentence} shows example of sentences where the text embedding alone does not classify accurately but is improved by other modalities (Figure \ref{fig:other_better}). On the left, the text alone classifies as change, on the right as sustain, 
when the true classification is neutral. It also shows an example where only text alone classifies the sentence correctly as change, and the model is not misled by other modalities (see Figure \ref{fig:text_better}).

\section{Conclusion and Future Work}
In this paper, we present a multimodal classifier for the three MISC classes of client behavior: change talk, sustain talk, and follow neutral. Our classifier is based on AnnoMI, an open access Motivational Interviewing database that is annotated in MISC classes and has been transcribed. We reorganized the transcript into sentences with lexical meaning and performed multimodal annotations of facial and body expressivity. Taking advantage of these multimodal inputs, we train a classifier that achieves greater accuracy than a unimodal approach and outperforms the existing approaches. We also use self-attention layers to determine the contribution of each modality, allowing us to interpret the results of our classifier and identify the most informative modality for a given sentence.

In future work, we plan to improve the model's performance by fine-tuning the Bert and Beats transformers. In addition, we envision endowing a virtual therapist agent with this model to enable it to detect whether the client is responding to therapy and is producing change talk. The agent could also provide feedback to the user regarding why it detected that the client may not be ready to change (e.g., tone of voice).
Finally, we aim to make the model publicly available to facilitate the annotation of new MI videos and serve as a baseline for future work. Overall, our approach demonstrates the value of multimodal input in improving the accuracy of MISC classification while providing interpretable features.

\section*{Acknowledgement}
  This work was partially funded by the ANR-DFG-JST Panorama and ANR-JST-CREST TAPAS (19-JSTS-0001-01) projects.

\bibliographystyle{ACM-Reference-Format}
\bibliography{sample-base}

\end{document}